\documentclass[a4paper]{article}

\usepackage{INTERSPEECH2021}
\usepackage{color}
\title{Self-supervised Dialogue Learning for Spoken Conversational Question Answering}

\name{Nuo Chen$^{1 \dagger}$\thanks{$^{\dagger}$ Indicates equal contribution}, Chenyu You$^{2 \dagger}$, Yuexian Zou$^{2,3,*}$\thanks{$^{*}$ Corresponding Author}}
\address{$^{1}$ADSPLAB, School of ECE, Peking University, Shenzhen, China\\
$^{2}$Department of Electrical Engineering, Yale University, CT, USA\\
$^3$Peng Cheng Laboratory, Shenzhen, China
}

\email{nuochen@pku.edu.cn, chenyu.you@yale.edu, zouyx@pku.edu.cn}

\begin{document}

\maketitle
\begin{abstract}
    In spoken conversational question answering (SCQA), the answer to the corresponding question is generated by retrieving and then analyzing a fixed spoken document, including multi-part conversations. Most SCQA systems have considered only retrieving information from ordered utterances. However, the sequential order of dialogue is important to build a robust spoken conversational question answering system, and the changes of utterances order may severely result in low-quality and incoherent corpora. To this end, we introduce a self-supervised learning approach, including \textit{incoherence discrimination}, \textit{insertion detection}, and \textit{question prediction}, to explicitly capture the coreference resolution and dialogue coherence among spoken documents. Specifically, we design a joint learning framework where the auxiliary self-supervised tasks can enable the pre-trained SCQA systems towards more coherent and meaningful spoken dialogue learning. We also utilize the proposed self-supervised learning tasks to capture intra-sentence coherence. Experimental results demonstrate that our proposed method provides more coherent, meaningful, and appropriate responses, yielding superior performance gains compared to the original pre-trained language models. Our method achieves state-of-the-art results on the Spoken-CoQA dataset. 
     
\end{abstract}
\noindent\textbf{Index Terms}: self-supervised learning, dialogue learning, spoken conversational question answering

\section{Introduction}

Spoken conversational question answering (SCQA) considers the problem of retrieving and aggregating spoken documents from a fixed collection, and then analyzes  retrieved information to provide answers according to the given questions \cite{you2020data}. In recent years, neural network based methods \cite{you2021momentum,you2019ct,you2020unsupervised,you2018structurally,you2019low} have attracted a lot of attention due to the advance in deep learning. Specifically, voice interfaces for dialogue systems have drawn massive attention of researchers. Several methods have been widely applied in various real-world scenarios, such as Microsoft XiaoIce and Apple Siri.

\begin{figure}[t]
  \centering
  \includegraphics[width=0.85\linewidth]{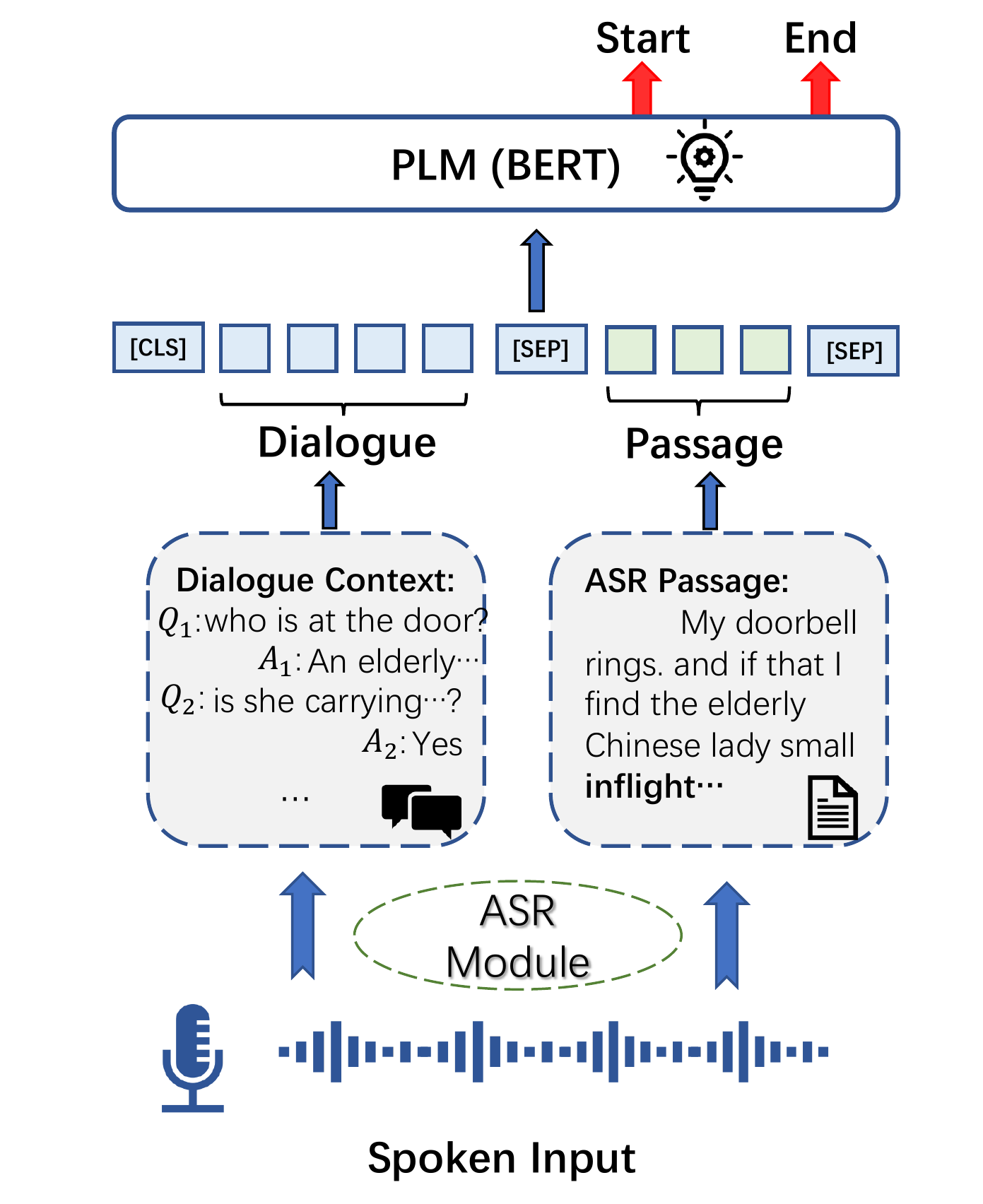}
  \vspace{-10pt}
  \caption{The overview of the PLMs-based SCQA pipeline. \textit{"start"} and \textit{"end"} indicate the start and end indexes  of prediction answers, respectively.}
  \label{fig:pipeline}
  \vspace{-20pt}
\end{figure}

Prior SCQA systems focus on utilizing pre-trained language models (PLMs) as the backbone models and achieve promising results \cite{you2020data,you2020knowledge,you2020contextualized}. However, building such systems still faces two major challenges: 1) almost any SCQA systems for spoken documents \cite{you2020knowledge,you2020contextualized,lee2018odsqa} are basically composed of two systems (Automatic speech recognition (ASR) system and CQA system) to tackle these SCQA tasks. First, they utilize an ASR module to translate spoken contents into the corresponding transcripts. Then a Text-CQA module adopts them as input, and  provides the final prediction answer. An example from SCQA procedure is depicted in Figure \ref{fig:pipeline}. In general, ASR transcripts are not well-applicable to every spoken document. Thus, in this procedure, the network performance is mainly limited by the highly noisy data in ASR transcripts due to  the word recognition errors (e.g., \textit{poor} to \textit{pool} ), and lack of capitalisation and punctuation. 2) The utterance order in a meaningful and coherent dialogue largely determines the response generation accuracy \cite{li2016diversity,wu2019self}. Existing SCQA systems rely heavily on contextual understanding in the dialogue flow. Thus, how to utilize the sequential order within spoken conversational documents as the self-supervisory signal to guide coherent dialogue learning is essential for the model developments.

\begin{figure*}[t]
  \centering
  \includegraphics[width=0.95\linewidth]{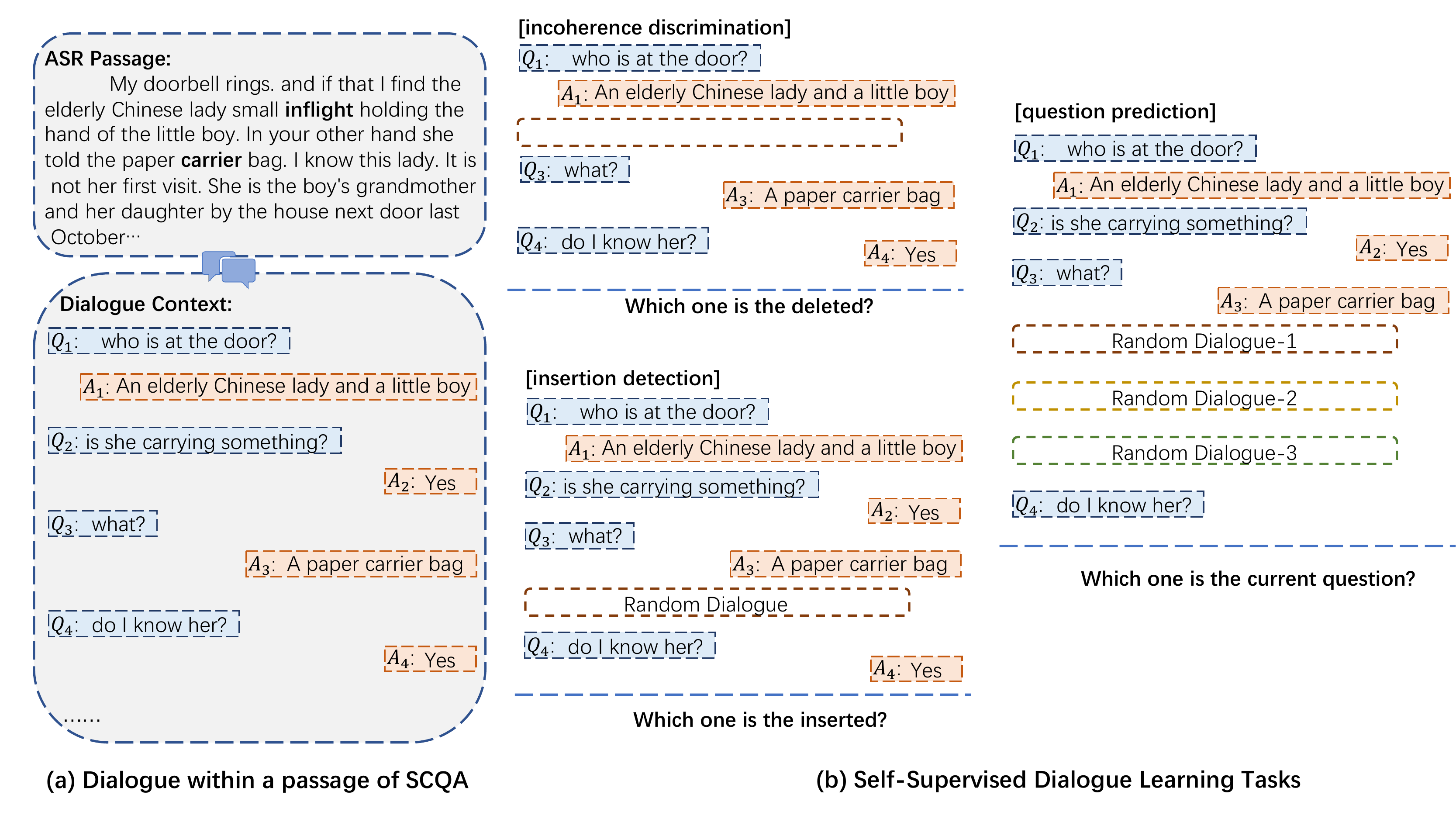}
  \vspace{-10pt}
  \caption{An overview of our proposed self-supervised dialogue learning framework. The input sequence for each sub-task ($k$ consecutive dialogue utterances) is dynamically selected from the original dialogue history within the fixed passage during training. For example, ``Random Dialogue" is the utterance that is randomly extracted from an random dialogue. }
  \label{fig:model}
  \vspace{-10pt}
\end{figure*}

To tackle the above-mentioned challenges, several studies have been proposed. Recent studies \cite{lee2018odsqa,li2018spoken} explored utilizing sub-word units to replace the whole word in order to reduce ASR errors. More recently, \cite{DBLP:conf/interspeech/KuoLC20} combined speech content and text transcriptions based on PLMs to mine potentially useful clues in acoustic signals for better answer predictions, which may be influenced negatively by highly noisy ASR transcriptions. Most recently, \cite{you2020data,you2020contextualized} introduced a teacher-student framework to distill incorruptible knowledge from a weighted teacher to make the student more ASR robust. However, none of existing works explicitly model the sequential order and evaluate its conversation criticality to the developments of SCQA systems.

In this paper, we work towards modeling dialogue flow within a fixed passage for the downstream SCQA tasks with performance improvements. The objective is achieved by exploring the coherence and consistency in a conversation as the useful self-supervised signals to guide spoken dialogue learning. In implementation, we introduce several self-supervised dialogue learning tasks to explicitly capture coreference resolution, semantic relevance, and dialogue coherence. Specifically, the auxiliary tasks include three sub-tasks: \textit{incoherence discrimination}, \textit{insertion detection}, and \textit{question prediction}. Then we jointly train a PLMs-based SCQA model in a multi-task manner to improve the model capability by learning task-specific knowledge to accurately predict answers while preserving dialogue coherence. Moreover, the proposed self-supervised tasks do not require additional annotation and thus can be widely applied in other downstream speech question answering tasks. Extensive experiments show that the proposed self-supervised dialogue learning tasks consistently bring remarkable performance improvements for state-of-the-art PLM-based SCQA models, and achieves superior results on Spoken-CoQA \cite{you2020data} dataset.

\section{Methods}
In this section, we first describe the task of spoken machine reading comprehension. We then introduce how to apply PLMs to trackle this task. In the end, we present the proposed self-supervised dialogue learning method on modeling the inconsistent order detection, as shown in Figure \ref{fig:model}.

\subsection{Problem Formulation }
\label{sec:probfomula}
In this study, we aim at engaging in conversation with humans in a dialogue-style interaction \cite{gao2018neural}. Given a SCQA dataset, we can formulate it as $\mathcal{D}$ = $\{P_{i},Q_{i},A_{i}\}_{1}^{N}$, where $P_{i}$ denotes the given passage. \!$Q_{i}$=$\{q_{i}^{1},q_{i}^{2},...,q_{i}^{L}\}$ and $ A_{i}$= $\{a_{i}^{1},a_{i}^{2},...,a_{i}^{l}\}$ are $l$-turn dialogue questions and the corresponding answers based on the $\{P_{i}\}_{1}^{N}$, respectively. For example, given the passage and the corresponding dialogue question $\{P_{i},q_{i}^{l}\}$, the task is to provide the answer $a_{i}^{l}$. The response depends on both the passage and history dialogue utterances. Inspired by recent success \cite{zhu2018sdnet}, we prepend the latest $m$ rounds of dialogue to the current question $q_{i}^{l}$ by incorporating conversation history for answer predictions. Consequently, the questions can be reformulated as $q_{i}^{l}$=
\{$q_{i}^{l-m},a_{i}^{l-m};..;q_{i}^{l-1},a_{i}^{l-1}\}$.

\subsection{Pre-trained Language Models}
Existing PLMs-based methods \cite{devlin2018bert,lan2019albert,yang2019xlnet,liu2019roberta} have achieved superior performance in a wide range of natural language processing tasks, such as question answering \cite{huang2018flowqa,seo2016bidirectional,huang2017fusionnet,chen2020adaptive} and text classification \cite{DBLP:conf/acl/ChenYY20}. In this work, we build upon their success and evaluate our proposed method by incorporating it into BERT \cite{devlin2018bert}, which is the widely used PLMs-based architecture.

\subsection{BERT for SCQA} 
We employ this PLMs-based approach in this work by applying BERT for SCAQ tasks. The input sequence is obtained by concatenating auto-transcribed token sequence of dialogue questions and corresponding passage. In general, the input is formulated as,
\begin{equation}
    \mathbf{X}=[\mathrm{[CLS] q_{i}^{1} a_{i}^{1} ... q_{i}^{l} a_{i}^{l}[SEP] P_{i} [SEP]}],
\end{equation}
where ``[CLS]" indicates the beginning token of every word sequence and ``[SEP]" is a special separator token. Then the pre-trained BERT is employed to extract hidden state features of each token. Next, a specific task layer is applied to predict an emphasis probability distribution over these representations. Finally, we adopt the cross-entropy loss as the training objective. In detail, suppose that the $k$-th token and $j$-th token in the $i$-th training examples are the start and end index of the answer span for $l$-th turn dialogue, SCQA answer prediction loss is given as:
\begin{equation}
    \mathcal{L}_{ans}=-\sum_{i}\sum_{l}^{}[log(p_{i,l}^{k,start})+log(p_{i,l}^{j,end})]	
\end{equation}

\subsection{Self-Supervised Dialogue Learning Tasks }
We aim at capturing dialogue coherent and consistency to produce more meaningful representations for performance improvements. Figure \ref{fig:model} illustrates the overview of the SCQA task and our proposed methods. Specifically, we design three auxiliary self-supervised tasks: \textit{incoherence discrimination}, \textit{insertion detection}, and \textit{question prediction}. These tasks are jointly trained with the SCQA model in a multi-task manner. More concretely, we introduce two special tokens $\mathrm{[INC],[INS]}$ for two following downstream tasks, respectively. 
 
\subsubsection{Incoherence Discrimination}
We argue that the sequential order of utterances within the dialogue is important for meaningful dialogue learning to solve the SCQA tasks. However, as noted in \cite{lan2019albert}, BERT has the limitation in learning discourse-level semantic information due to NSP (next sentence prediction) \cite{wu2019self}. In the context of SCQA, the model is required not only to discriminate whether the utterances are consecutive or not even though they are semantically related words, but also to distinguish the utterances with different semantic meanings. 

To achieve this goal, we design \textit{incoherence discrimination}. Specifically, we first extract $k$ consecutive utterances from the dialogue context with respect to $i$-th passage, and then randomly delete one of them. In the training stage, we add $\mathrm{[INC]}$ tokens before each utterance to help the model find which is the deleted utterance. $\mathrm{[INC]}$ tokens refer the possible position of the deleted target utterance. Hence, the input for this sub-task can be given as:
 \begin{equation}
 \begin{split}
     \mathbf{X_{INC}}=[\mathrm{[CLS][INC]_{1}u_{i}^{1}[INC]_{2}\cdot\cdot\cdot u_{i}^{t-1}[INC]_{t}}\\
     \mathrm{u_{i}^{t+1}\cdot\cdot\cdot[INC]_{k}u_{i}^{k}[SEP]u_{i}^{t}[SEP]}],
     \end{split}
 \end{equation}
where $u_{i}^{k}$ denotes vector concatenation of $q_{i}^{k},a_{i}^{k}$, and $u_{i}^{t}$ is the deleted utterance.

\subsubsection{Insertion Detection}

Existing BERT-based models for multi-turn conversations have achieved superior performance in capturing discourse-level representations within the entire dialog history. However, these model in multi-turn manner only learn token-level  information \cite{lan2019albert}, ignoring utterance-level interaction. 

To enrich the utterance-level interaction between the utterances within a dialogue, we propose \textit{insertion detection}. Formally, we first extract $k$ consecutive utterances from $i$-th passage of the original dialogue contexts. Then we insert an utterance to a random dialogue from a set of  $k$ extracted utterances. For example, $k+1$ utterances consist of $k$ utterances from original spoken documents and one from different corpus. The objective is to detect which one is the inserted among the k+1 utterances. $\mathrm{[INS]}$ tokens are positioned before each utterance. The input is given as follows:

\begin{equation}
 \begin{split}
      \mathbf{X_{INS}}=[\mathrm{[CLS][INS]_{1}u_{i}^{1}[INS]_{2} \cdot\cdot\cdot [INS]_{t}u_{i}^{ins}}\\
      \mathrm{[INS]_{t+1}u_{i}^{t}\cdot\cdot\cdot[INS]_{k+1}u_{i}^{k}[SEP]}]
 \end{split}
 \end{equation}
where $u_{i}^{ins}$ is the inserted utterance.

\subsubsection{Question Prediction}
In contrast to two previous auxiliary self-supervised tasks designed to capture dialogue consistency and coherence within a properly ordered dialog, we propose \textit{question prediction} to learn temporal dependencies between semantically similar utterances within a dialogue. Given a dialogue history, $C_i=\{u_{i}^{1},u_{i}^{2},..,u_{i}^{t-1}\}$, randomly shuffled $k$ questions, $Q=\{q_1^{rand},q_2^{rand},u_{i}^{t},..,q_k^{rand}\}$ and the passage $P_i$, $u_{i}^{t}$ denotes the current corresponding question, the \textit{question prediction} in this process aims to predict which question comes from a given context. Following the multi-choice question answering setting \cite{DBLP:conf/interspeech/KuoLC20,DBLP:conf/acl/YanZJZ20}, the input can be written as:
\begin{equation}
    \mathbf{X_{QUE}}=[\mathrm{[CLS]C_i,Q_m[SEP]P_i[SEP}]
 \end{equation}
where $Q_m$ is the $m$-th question among $Q$. In this process, to select a question candidate, we use a signal layer \cite{DBLP:journals/corr/abs-2009-04703} to translate the hidden representation of "[CLS]" for similarity scores. We then stack all the scores of $k$ question candidates as input signal, and pass them to the softmax function. In this process, the selected tokens for each task (e.g., $\mathrm{[INS]_t,[INC]_t}$) are labeled as 1 or 0. We utilize binary cross-entropy loss as the optimization objective for these self-supervised tasks. Finally, the overall loss function is computed by combining SCQA answer prediction loss $\mathcal{L}_{ans}$ and auxiliary tasks loss with same ratio.

\begin{table*}[t]
\caption{Experimental results on the Spoken-CoQA test set. We use BERT-base as the baseline in this analysis. INC, INS, and QUE denote that the model trained with incoherence discrimination, insertion detection, and question prediction, respectively.}
  \vspace{-5pt}
    \centering
    \begin{tabular}{l|c c c c c |c}
    \toprule
   \textbf{Methods} &\textbf{Child.}&\textbf{Liter.}&\textbf{Mid-High.}&\textbf{News}&\textbf{Wiki}&\textbf{Overall} \\
    \hline
    \hline
      FlowQA \cite{huang2018flowqa}  & 35.8&35.2&34.2&33.6&34.7&34.7\\
      BERT &55.6&54.6&52.7&53.8& 53.8&54.1\\
      BERT  + SLU \cite{serdyuk2018towards}  & 55.7&54.6&53.1& 54.0& 54.6&54.4\\
      BERT  + Sub-Word \cite{li2018spoken}&56.6&56.7&53.9&55.0&54.8&55.4\\
      BERT + Cross-Attention \cite{you2020contextualized}
      & 57.1&56.1&56.0& 54.6&55.0& 55.6\\
        \hline
        \hline
    BERT + INC &58.2&57.7&55.6&55.7&56.3&56.7\\
    BERT + INS &57.2&56.7&55.6&55.2&55.3&56.0\\
    BERT + QUE &58.0&57.4&55.6&55.4&55.6&56.4\\
     BERT + INS + QUE&57.4&56.6&55.0&56.4&55.7&56.7\\
     BERT+ INC + QUE &58.1&57.8&56.3&56.8&56.2&57.2\\
     BERT + INC+ INS&58.2&57.6&55.4&56.0&56.1&56.9\\
     \hline
       \textbf{BERT + INC + INS + QUE (Ours)}  &\textbf{59.3}&\textbf{58.4}&\textbf{56.8}&\textbf{57.6}&\textbf{56.8}&\textbf{57.8}\\
      
        \toprule
    \end{tabular}
  \vspace{-15pt}
    \label{table:domain}
\end{table*}

\section{Experiments}

\subsection{Dataset}
We evaluate our proposed method on Spoken-CoQA dataset. Spoken-CoQA~\cite{you2020data} is an English spoken conversational question answering (SCQA) dataset, which consists of $40$k question-answer pairs from $4$k conversations in the training set and $380$ conversations in the test set from five diverse domains, respectively. In Spoken-CoQA, questions are required to understand both passages and previous dialogue history, which is a challenging task to SCQA models. Note that questions and passages are both in text and spoken form, and answers are in text form, respectively. The word error rate (WER) is 18.7$\%$.

\subsection{Experimental Settings}
We implement our model by using PyTorch. We adopt BERT-based as our backbone encoder, which consists of $12$ transformer layers with maximum sequence length of the input set to $512$ and the hidden vector dimension $768$. The $m$ and $k$ are set to 2 and 9, respectively. In the training stage, the learning rate is set to $3\times 10^{-5}$ and batch size is 4 per GPU. We train each model on 2x 2080Ti for 2-3 days with 6 epochs. We utilize F1 and Exact Match (EM) for evaluating SCQA quality.

\subsection{Results}

We compare our proposed methods with five baselines\footnote{For fair comparison, we do not compare our methods with baselines that have not been published.} in Table \ref{table:domain}. We observe that our model significantly outperforms other baselines in most of the cases on Spoken-CoQA. This indicates that the proposed self-supervised learning tasks can explicitly enable the model to capture coherence and consistency within a dialogue. Compared with BERT, our method improves F1 score to 57.8$\%$ (vs.54.1$\%$). We also conduct ablation studies and report the results in Table \ref{table:domain}. We see that sequentially add three proposed strategies bring significant performance improvements in F1 score. The results show that INC $>$ QUE $>$ INS, with respect to the importance of auxiliary manipulation strategy. This indicates that incorporating sequential information with the dialogue help to learn more contextual representations from the sequences. We also see that removing INS leads to the small performance drop (57.0$\%$ vs. 57.8$\%$). This is because inserting a random question to some extent does not affect the discourse relations in the multi-turn conversations. Removing QUE, the F1 score for SCQA systems is dropped from 57.8$\%$ to 56.6$\%$. This suggests that, by predicting which question comes from a given dialogue history and passage, the network can facilitate more interactions between passage and question for superior performance boosts.
 
\begin{table}[t]
\caption{Experimental results on Spoken-CoQA dataset when prepending different number of history questions and answers to the current question.}
  \vspace{-5pt}
    \centering

    \begin{tabular}{c c c}
    \toprule
      Dialogue History rounds $m$   & EM&F1  \\
      \hline
      0&34.2&44.2\\
      1&39.8&52.0\\
       2  & \textbf{40.6}&\textbf{54.1}\\
       3& 40.0&53.1\\
       4&39.3& 51.2\\
       \bottomrule
    \end{tabular}
  \vspace{-15pt}
    \label{tab:my_label}
\end{table}

\section{Discussion}
\subsection{Dialogue Context History}
In this study, we utilize dialogue context history via prepending the latest $m$ rounds of dialogue with the current question to incorporate contextual information, including questions and ground-truth answers (See Section \ref{sec:probfomula}). Here we investigate the effect of $m$, and report the results in Table \ref{tab:my_label}. Specifically, we utilize BERT as the baseline. We observe that excluding dialogue history results in performance degradations by 5.6$\%$ and 7.8$\%$ in terms of EM and F1 score. This suggests that incorporating dialogue contextual information brings significant performance improvements. Meanwhile, we can see that when $m$ is set to $2$, the model achieves the best performance.

\begin{figure}
    \centering
    \includegraphics[width=0.6\linewidth]{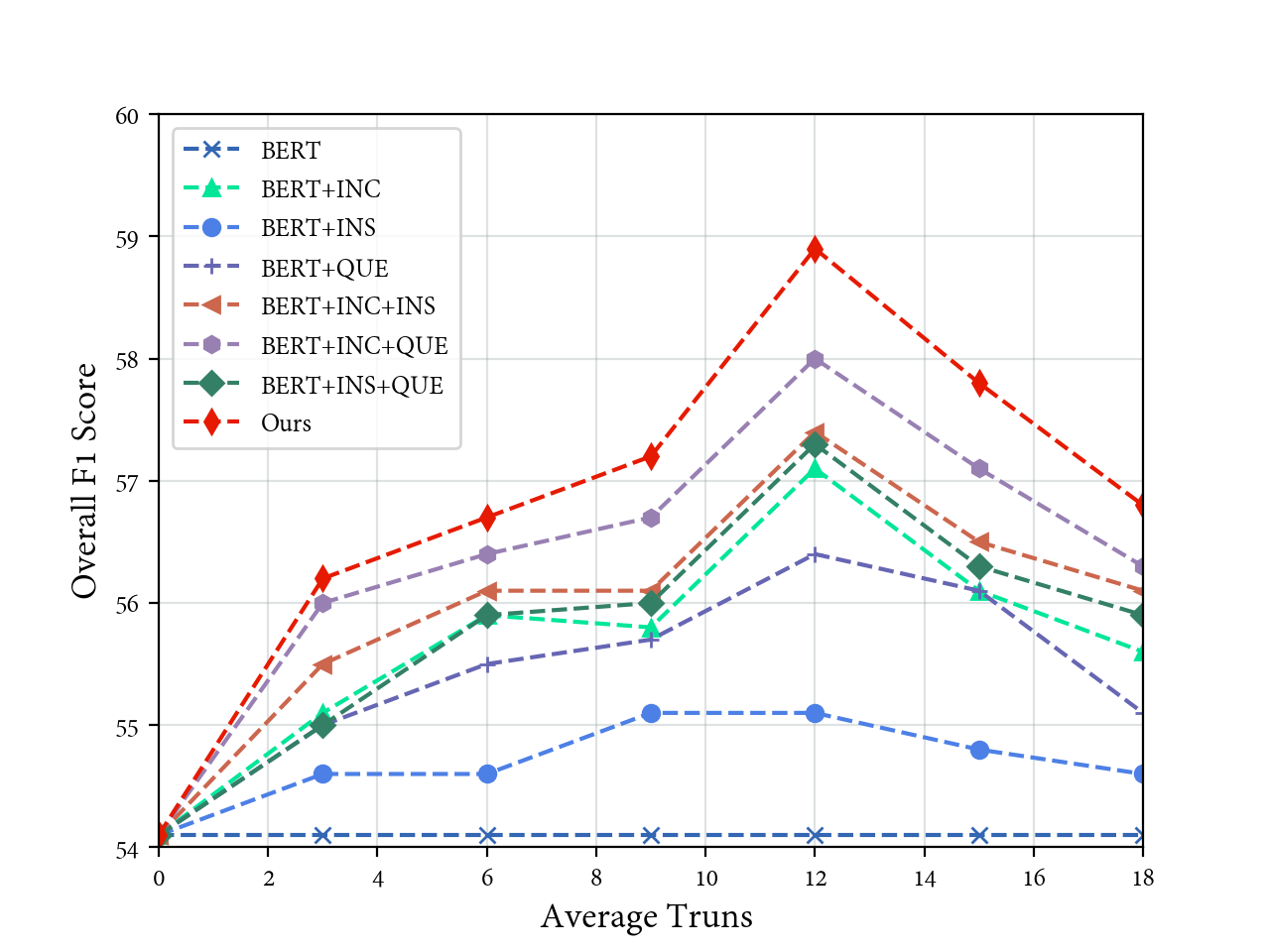}
  \vspace{-10pt}
    \caption{Experimental results of three auxiliary self-supervised tasks with different lengths of dialogue turns on Spoken-CoQA test set.}
    \label{fig:all}
  \vspace{-15pt}
\end{figure}

\subsection{Different Lengths of Dialogue Context}
We conduct experiments to explore the impact of different lengths of dialogue turns on network performance. We report the experimental results of BERT with three auxiliary tasks by examining all possible combinations in Figure \ref{fig:all}. We can observe that the performance of training with three proposed self-supervised tasks shows a tendency to increase significantly ($\leq 12$). In addition, there exists the performance drop when the number of turns keeps increasing ($\geq 12$). This is because when trained with a few dialogues, the model cannot distinguish dialog consistency and coherence. On the other hand, when the number of average turns increase, rich dialogue history may introduce additional noisy signals, which significantly degrade the performance of the SCQA model. Based on the observations of training with three manipulation strategies, all of them show absolute improvements compared with the baseline. We achieve the best results when all the manipulation strategies are trained simultaneously (See red line in the Figure \ref{fig:all}).

\section{Conclusion}
In this paper, we introduce a self-supervised dialogue learning for spoken conversational question answering by explicitly modeling dialogue coherence and consistency. Specifically, the model is fully trained in multi-task self-supervised manner and can be easily applied into existing state-of-the-art methods. Experimental results show that proposed methods achieve  state-of-the-art performance on Spoken-CoQA dataset. In future, we expect our proposed self-supervised tasks in other language tasks, leading to further performance gains.

\section{Acknowledgements}
This paper is partially supported by Shenzhen municipal research funding project and Technology Fundamental Research Programs (No: GXWD20201231165807007-20200814115301001 \& JSGG20191129105421211).

\bibliographystyle{IEEEtran}

\bibliography{mybib}

\end{document}